\documentclass{llncs}
\usepackage{graphicx} 
\usepackage{xcolor}
\usepackage{amsmath}
\usepackage[misc]{ifsym}
\usepackage{algorithm2e}
\DontPrintSemicolon
\usepackage{wrapfig}
\usepackage{listings}
\usepackage{fancyhdr}
\begin{document}

\title{Multi-Objective Coevolution of Prompts and Templates for Circuit Approximation}

\author{Martin Tomasovic \and
Lukas Sekanina\orcidID{0000-0002-2693-9011}}
\institute{Brno University of Technology, Faculty of Information Technology\\Brno, Czech Republic\\
\email{xtomas36@stud.fit.vut.cz, sekanina@fit.vut.cz}}%

\maketitle
\thispagestyle{fancy}

\begin{abstract}
Approximate multipliers deliberately relax computational accuracy to achieve gains in power efficiency, latency, and silicon area, which makes them well-suited for error-resilient applications such as neural networks. In this work, we introduce a co-evolutionary algorithm that leverages an off-the-shelf large language model (LLM) without requiring domain-specific training to automate the design of optimized 8-bit approximate multipliers. The approach simultaneously evolves a population of candidate circuits and a population of prompt templates that steer LLM-driven modifications. Experimental results for several target design objectives demonstrate that the proposed method discovers approximate multipliers with improved error–area trade-offs compared to highly optimized circuits from the EvoApproxLib library. 
\end{abstract}

\section{Introduction}

Approximate circuits are digital circuits intentionally designed to operate inexactly~\cite{Jiang:axc:surv:2020,ACsurvey:ACM:2020}. Compared to precise implementations, they offer lower power consumption, reduced latency, or smaller chip area. These advantages are particularly beneficial in error-resilient applications, such as deep neural networks, where approximate implementations can achieve reduced power consumption and latency at the cost of an acceptable level of output error~\cite{ArmeniakosZSH23}.

An important and intensively studied approximate circuit is the approximate multiplier, as multiplication is one of the most frequently executed operations during both the training and inference phases of neural networks. In addition to conventional design approaches~\cite{JiangLL019:bookch}, evolutionary design methods have been proposed to design approximate multipliers and have demonstrated particularly strong performance in this domain~\cite{mrazek:date:17,ceska:iccad17,Mrazek:2020:approxMultipliersForCNN}. The approximation process usually begins with a gate-level multiplier, and the task for evolution is to modify it to obtain implementations that exhibit desired error-area-delay tradeoffs. The EvoApproxLib library provides a collection of approximate arithmetic circuits designed using Cartesian genetic programming (CGP) and is widely adopted in both practical applications and research studies~\cite{mrazek:date:17}. As circuits in EvoApproxLib have been obtained by time-consuming, highly optimized search and verification techniques, it is challenging to find their improved implementations. 

We are aware of only one work dealing with approximate circuit design using large language models (LLMs). Yi et al. introduced a GPTAC method -- a domain-specific generative pre-trained model for approximate circuit design~\cite{Yi:GPTAC:2024}. The model can generate approximate circuits with specific area and accuracy requirements from a large dataset of learned circuit designs. Trained LLM is used in the design loop; however, the design process does not show properties of an evolutionary algorithm (EA). 

In this paper, we show that a common LLM (without any additional training on a dataset of approximate circuits) can generate competitive approximate multipliers if properly prompted. Based on several preliminary experiments with LLMs and their integration with an EA, we propose a co-evolutionary EA (CoEA) operating with two populations: (1) a population of circuits, represented using a set of boolean expressions, in which LLM serves as a template-based modification operator for creating new candidate circuits from existing ones; (2) a population of templates specifying how to prompt LLM to generate well-performing circuits. Candidate templates are evaluated using a subset of circuits from the population of circuits. Templates are modified using a set of template-modifying operators based on LLM. Circuits are evaluated using two metrics: approximation error and chip area. A multi-objective selection scheme is employed to create a new population of candidate circuits. The paper makes the following key contributions: 
\begin{itemize}
\item We develop a first multi-objective co-evolutionary system for circuit approximation based on LLM, which does not require any pretraining on a set of approximate circuits.
\item For 8-bit approximate multipliers we show that the proposed co-evolutionary method can create circuits showing better trade-offs than the highly optimized circuits available in the state-of-the-art EvoApproxLib library of approximate circuits.
\end{itemize}

\section{Related Research}

\subsection{Approximate circuits}

As outlined in the survey~\cite{Jiang:axc:surv:2020}, approximate arithmetic circuits can be realized across three main abstraction levels: (1) At the \emph{algorithm level}, the standard exact multiplication is substituted with an alternative method that delivers approximate but acceptably accurate results, for instance, a logarithm-based multiplier (ALM)~\cite{ALM}.
(2) At the \emph{architecture level}, a common circuit structure is initially selected and then altered at different stages, such as partial product generation, accumulation, or compression—based on manual design decisions or heuristic approaches. For example, in a truncation-based multiplier, the least significant partial products are intentionally omitted, reducing hardware complexity and power consumption at the cost of a controlled loss in accuracy.
(3) At the \emph{circuit level}, approximation techniques are applied directly at the gate level, for example, through circuit rewriting~\cite{mrazek:date:17,Yi:GPTAC:2024}, to produce more compact and simplified circuit designs.

Except for the GPTAC method that utilized a LLM trained using a large collection of approximate circuits~\cite{Yi:GPTAC:2024}, LLMs have not been used for approximate circuit design. However, their utilization in electronic design automation has been rapidly growing because hardware designs and intermediate scripts can be represented as text. They have initially been used as Verilog code generators~\cite{VeriGen:2024}; currently, their use spans from logic and high-level synthesis via processor and accelerator design to testing and verification~\cite{Xu:socc2025}.

\subsection{Evolutionary design methods}

Cartesian Genetic Programming (CGP) has been applied to synthesize approximate arithmetic circuits. The methodology, detailed in~\cite{mrazek:date:17,ceska:iccad17,Mrazek:2020:approxMultipliersForCNN}, starts from an exact gate-level implementation chosen by the user, which is mapped on the cartesian grid of computational nodes of CGP. A mutation-based search algorithm tries to deliver implementations showing excellent trade-offs between the error and hardware metrics. The error introduced by an approximate arithmetic circuit can be evaluated using various metrics, including the \emph{worst-case error} (WCE) given in Eq.~\ref{eq:WCE} and the \emph{mean squared error} (MSE) defined in Eq.~\ref{eq:MSE}.

\begin{equation}
\text{WCE} = \max_{\forall i} \left| O_{\text{approx}}^{(i)} - O_{\text{orig}}^{(i)} \right|
\label{eq:WCE}
\end{equation}

\begin{equation}
\mathrm{MSE} = \frac{\sum_{\forall i}  (O_{\text{approx}}^{(i)} - O_{\text{orig}}^{(i)})^2} {2^{n}},
\label{eq:MSE}
\end{equation}
where $O_{\text{orig}}^{(i)}$ represents the exact output, while $O_{\text{approx}}^{(i)}$ denotes the approximate output (expressed in decimal form) for a specific input $i$ of an $n$-bit circuit (with $n = 2 \cdot b$ in the case of a $b$-bit multiplier). For smaller circuits, the error is obtained using circuit simulation for all possible input combinations. For larger circuits, either formal methods are employed~\cite{ceska:iccad17} or the error is estimated.  
Hardware characteristics such as area, delay, and power consumption are typically estimated to speed up the design exploration (see Section~\ref{sec:setup}).  The EvoApproxLib\footnote{\url{https://ehw.fit.vutbr.cz/evoapproxlib/}} library of approximate circuits is composed of approximate circuits produced in a large number of computationally expensive CGP runs~\cite{Mrazek:2020:approxMultipliersForCNN}. Further improvements of these circuits is both challenging and computationally expensive. 

\emph{Co-evolutionary algorithms} (CoEAs) are defined by how they assess the fitness of individuals within the system. In these approaches, an individual’s fitness is not evaluated in isolation, but rather depends on its interactions with members of one or more other populations~\cite{Popovicci:Coevolutionary}. 
CoEAs can be broadly categorized based on the nature of the solution they target. Two main problem types addressed by CoEAs are usually characterized as: \emph{compositional} (cooperative) problems (e.g., \cite{potter2000cooperative}) and \emph{test-based} problems (e.g., in circuit design, co-evolving the training data and circuits in~\cite{Hrbacek:2013:ECAL}). However, CoEAs have not been applied in the evolutionary design of approximate circuits yet.

\subsection{LLMs in evolutionary computation}

The combination of large language models (LLMs) with evolutionary algorithms (EAs) has recently attracted considerable attention, especially for program synthesis~\cite{LiventsevGHM23}, as reflected in several survey studies~\cite{Wu:LLM:EA:2025,Hemberg2025,Sobania:tec:2025}. In particular, Hemberg et al.~\cite{Hemberg2025} categorize the interaction between genetic programming (GP) and LLMs into three main groups: (i) LLMs integrated directly into GP, for example, as advanced mutation operators~\cite{shem:2025:evoTrans}; (ii) LLMs used to support GP, such as for test case generation or representation design~\cite{CaetanoTP23}; and (iii) GP applied to improve LLM-related tasks, for instance, through prompt optimization.

A representative example of the first category is the work of Shem-Tov et al.~\cite{shem:2025:evoTrans}, who utilize the BERT Transformer as a context-aware mutation operator in GP. Their approach leverages both the structural information of program trees and historical fitness data to guide node replacement, leading to improved convergence behavior. In a related direction, LLMs have also been used to refine grammars for program synthesis within grammatical evolution frameworks~\cite{Zarb:lmm:ge:2025}.

Another line of research focuses on using LLM-inspired models as surrogates. Teixeira and Pappa~\cite{TexPappa:gecco25} introduce a transformer-based encoder that compares pairs of candidate solutions and predicts their relative quality, thereby reducing the need for explicit fitness evaluation at each generation. Xie et al.~\cite{Xie:KDD25} propose a co-evolutionary approach in which LLMs and configuration strategies are evolved jointly to improve the performance of surrogate-assisted evolutionary algorithms.

\section{Proposed Method}

The goal is to develop a method based on EAs and LLM for the automated design of approximate unsigned 8-bit multipliers. To avoid time-consuming training of LLM using a collection of approximate circuits, an off-the-shelf LLM is employed. 
The user is asked to provide target approximation parameters (e.g., WCE and area) and a few examples (up to 10) of approximate 8-bit multipliers with their errors and areas. After initial experiments with one-shot and few-shot prompts, we introduced a simple hill-climbing search algorithm to create useful prompts and thus generate circuits with desired behavior solely by LLM prompts; however, this approach has not led to competitive results either.

The circuit approximation problem is specific in the context of program synthesis using LLM for several reasons. Compared with common programming languages, hardware description languages are less represented in code repositories (such as GitHub) and consequently in LLM training datasets~\cite{HongRAURS:25}. When a candidate circuit showing unsatisfactory properties is created, the user is not usually able to ask LLM to correct the circuit's behavior for particular test case(s) on which the circuit had failed. The prompts have to be carefully optimized with respect to the circuit's global behavior, which is difficult. Hence, we introduce another evolutionary loop in which prompt templates are co-evolved with circuits.

\subsection{Coevolutionary approach}

The objective is to generate an approximate multiplier that closely matches the user-provided target error $T_E$ and area $T_A$ (this approach is typical in real applications). As the exact target values are usually not achievable, the algorithm is constructed to solve a multi-objective optimization problem; it builds a set of multipliers with error and area on the Pareto front near ($T_E$, $T_A$).

\begin{figure} 
    \centering
    \includegraphics[width=0.99\textwidth]{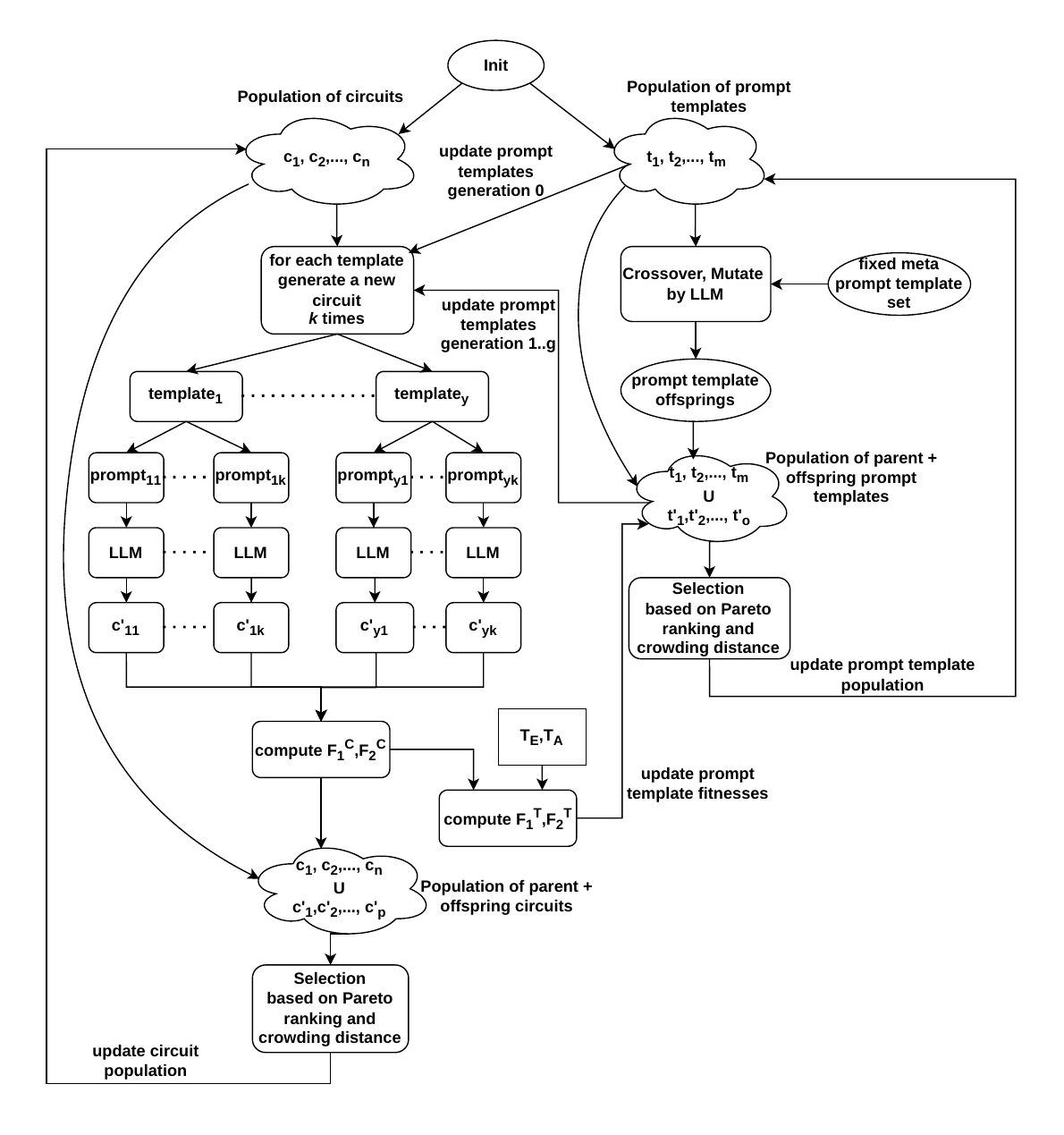}  
    \caption{The proposed circuit approximation method that utilizes a co-evolutionary algorithm evolving a population of circuits (left-hand side) and a population of circuit- producing templates (right-hand side).}
    \label{fig:method}
\end{figure}

Fig.~\ref{fig:method} shows the scheme of the method whose main idea is described in this section. Details about its components are provided in the following subsections. The method co-evolves two populations: (1) a population of circuits and (2) a population of circuit-producing templates (templates, for short). In the case of the evolution of circuits, LLM is used as a mutation operator, which generates a new candidate circuit from an existing one. New circuits are created using templates that are cultivated in the second population. Templates help to create strong prompts for LLM-based circuit modification. Their adaptation is essential as it is not always clear how to instruct LLM to build a useful offspring circuit. 

\subsection{Evolution of circuits}
\label{sec:meth:circ}

\emph{Circuit representation:} A candidate multiplier with two 8-bit operands INA and INB, and a 16-bit output OUT is represented as a list of logic expressions, e.g.:

\emph{Example Circuit 1:}
\begin{lstlisting}[breaklines=true, basicstyle=\ttfamily\small, backgroundcolor=\color{gray!10}]
N0 = INB7 & INA0
N1 = INB6 & INA1
N3 = N0 | N1
N4 = N0 & N1
N5 = INB5 & INA2
...
N117 = (N105 & N103) | (N116 & N115)
...
OUT15 = ((INA7 & N106) | (N118 & N117))
OUT14 = (N118 ^ N117)
...
OUT2 = 0
OUT1 = N88
\end{lstlisting}
The meaning of symbols is intuitive (and never explained to LLM); for example, INAj denotes the j-th bit of operand A, Nj denotes the j-th inner node, and OUTj denotes the j-th bit of the output product. Logic operations are expressed as in the C language. Constants (0 and 1) and parentheses can be utilized. A typical approximate multiplier can have hundreds of lines of code. We used this representation because it is similar to the logic equations on which LLMs are usually trained. The exact fine-tuning data set for the selected Qwen3-Coder-480B model is not available, but according to the technical report for the Qwen3 model, the data set used in the pretraining process contains mathematical and code-related data~\cite{qwen3technicalreport}.

\emph{Initialization:} The initial population is seeded with approximate 8-bit multipliers of various quality. 

\emph{Fitness functions:} Fitness $F^{C}_{1}(c)$ is the error of a candidate circuit $c$ determined according to Eq.~\ref{eq:WCE} or~\ref{eq:MSE} (depending on the objectives) for all  $2^{16}$ input vectors.
In addition, it is strictly requested that circuit $c$ performs the exact multiplication if at least one of the operands is zero. If this condition does not hold, then $F^{C}_{1}(c)=\infty$ is assigned to candidate circuit $c$. This requirement is critical for efficient deployment of approximate multipliers in DNN accelerators~\cite{Mrazek:2020:approxMultipliersForCNN}.
The second fitness, $F^{C}_{2}(c)$, is the area of $c$ on a chip, which is estimated with respect to the target fabrication technology as the sum of sizes of all gates involved in circuit $c$. The fitness values $F^{C}_{1}$ and $F^{C}_{2}$ are stored alongside the circuit code.

\emph{LLM-based Operators:} A new candidate circuit is generated by LLM using a prompt created according a template, e.g.: 

\emph{Example Template 1:}
\begin{lstlisting}[breaklines=true, basicstyle=\ttfamily\small, backgroundcolor=\color{gray!10}]
Given a circuit with these parameters: 
Worst case absolute error (WCE) = {approx_circuit_wce}
Area = {approx_circuit_area} um^2
The circuit: {approx_circuit}
Generate a new 8-bit unsigned approximate multiplier circuit with parameters:
Worst case absolute error (WCE) = {target_wce}
Area = {target_area} um^2
Generate only the new circuit, without any comments.
The new circuit:
\end{lstlisting}

Templates contain placeholders determined by \{...\}
for circuit parameters (errors, area) and the circuit description itself (a particular circuit description is given in Example Circuit 1). A template can refer to various existing circuits of the current population using the following identifiers: \texttt{random}, \texttt{closest to target}, \texttt{second closest to target}, a circuit with the \texttt{best error metric}, or a circuit with \texttt{the best area}. Circuits corresponding to these identifiers are stored and updated in local memory and can be used during the co-evolution. In a given generation, the algorithm employs up to $y$ templates, which can be reused. For a given circuit $c_i$ and a selected template, LLM is executed $k$ times to create $k$ offspring circuits (see the left loop of Fig.~\ref{fig:method}).

\emph{Selection:} The sets of parents and offspring are united, and from this union the selection algorithm selects the new population. The selection is implemented using the \texttt{RankAndCrowdingSurvival} function from \emph{pymoo} library and employs Pareto front ranking and crowding distance heuristics from NSGA-II~\cite{nsga2}.

\subsection{Evolution of templates}
\label{sec:meth:templ}

The evolution of templates is visualized in the right-hand loop of Fig.~\ref{fig:method}.

\emph{Representation:} A template is a blueprint for prompt creation. It contains immutable text and placeholders as illustrated in Example Template 1. The placeholders are dynamically replaced with specific data.
The following templates, stored in an external JSON file, can be utilized: Generate\_prompts, Mutation\_prompts, Crossover\_prompts, Post\_process\_prompt\_circuit and Post\_process\_
prompt\_prompt.

\emph{Initialization:} The initial population of templates contains manually-created templates for creating prompts. These prompts are used to instruct LLM to generate new candidate circuits in the first generation of circuit evolution.

\emph{Evaluation:} Fitness $F^{T}_{1}(t)$ of a template $t$ is defined as the ratio of valid and invalid candidate circuits generated by LLM using $t$ during one iteration of the circuit evolution loop. 
Fitness $F^{T}_{2}(t)$ of a template $t$ is defined as the average absolute difference between target error $T_E$ and the errors of candidate circuits generated using template $t$. Fitness values $F^{T}_{1}(t)$ and $F^{T}_{2}(t)$ are stored with each template $t$.

\emph{Template modifying operators:} 
A set of prompts is specified in advance for the mutation and crossover of templates; a given prompt is randomly selected when the genetic operation is carried out. The prompts contain a fixed text with instructions for LLM, variables like \texttt{\{valid\_count\}} denoting the number of valid circuits generated using this template, and particular template(s) for generating circuits. Examples of prompts follow:

\emph{Example prompt for mutation:}
\begin{lstlisting}[breaklines=true, basicstyle=\ttfamily\small, backgroundcolor=\color{gray!10}]
You are given a prompt template: "{prompt_template}"
This template generated {valid_count} valid 
and {invalid_count} invalid circuits. 
Modify the template, the new template should steer the large language model to generate circuits following more precisely given target parameters. Generate only the new template, without any comments. Output the new template:
\end{lstlisting}

\emph{Example prompt for crossover:}
\begin{lstlisting}[breaklines=true, basicstyle=\ttfamily\small, backgroundcolor=\color{gray!10}]
You are given two prompt templates
Template 1 with {valid_count1} valid and {invalid_count1} invalid circuits: "{prompt_template1}"
Template 2 with {valid_count2} valid and {invalid_count2} invalid circuits: "{prompt_template2}"
Create a new prompt template by combining the two prompt templates. The new templates should steer large language models to generate circuits with a smaller area while keeping WCE the same or lower. Generate only the new template, without any comments. Output the new template:
\end{lstlisting}

\emph{Selection:} The sets of parental and offspring templates are united, and from this union the \texttt{RankAndCrowdingSurvival} function from \emph{pymoo} library composes the new population.

\subsection{Co-evolutionary mechanisms}

The two evolutionary loops are synchronized every generation. For each template $t_i$, fitness $F^{T}_{1}(t_i)$ and $F^{T}_{2}(t_i)$ are computed after evaluation all circuits using $F^{C}_{1}$ and $F^{C}_{2}$. The co-evolution is terminated after $g$ generations. The co-evolutionary algorithm produces a set of circuits occupying a Pareto front (error vs. area) close to target $T_E$ and $T_A$.

\section{Setup}
\label{sec:setup}

For circuit design, we used Qwen3-Coder-480B, which is a mixture-of-experts coding model with 480B total parameters (approx. 35B active per inference), optimized for long-context (up to 256k tokens) code generation, reasoning, and agent-style software engineering tasks. For prompt templates, we used a smaller model, GPT-OSS-120B, which is a 120B-parameter open-weight LLM optimized for general reasoning and coding; temperature is $0.4$. Our software is implemented as a Python code that calls external LLMs. When circuits from EvoApproxLib are used, their Verilog code is transformed into the representation specified in Section~\ref{sec:meth:circ}. The error of a candidate circuit is obtained via simulation implemented using a highly optimized C code with OpenMP parallelization. 

We report the default settings for all parameters of the proposed method that were determined during experimentation with the algorithm and considering available computing resources (the number of LLM calls that we can perform). More detailed analysis of selected settings will follow in Section~\ref{sec:res:comp}. 

For the evolution of circuits, the settings are: Population size $n = 30$, repetition of LLM calls per template $k = 3$, max. templates $y = 3m$.
Circuit's fitness values $F^{C}_{1}$ and $F^{C}_{2}$ are computed as specified in Section~\ref{sec:meth:circ}. The circuit area is estimated using the gate sizes reported in Table~\ref{tab:gates}.

For template evolution, the settings are: Population size $m = 8$, mutation probability $= 0.9$, crossover probability $= 0.9$, and the number of prompt templates is 3 for mutation and 3 for crossover. The template fitness values $F^{T}_{1}$ and $F^{T}_{2}$ are computed as specified in Section~\ref{sec:meth:templ}. Each circuit/template iteration is repeated $g = 27$ times within a single-coevolutionary run. 

\begin{table}[]
\centering
\caption{Sizes of the gates in $\mu m^2$ corresponding to the 45~nm technology.}
\label{tab:gates}
\setlength{\tabcolsep}{5pt}
\begin{tabular}{l|l|l|l|l|l|l|l}
\textbf{Gate} & INV & AND & OR & XOR & NAND & NOR & XNOR \\ \hline
\textbf{Size} & 1.40 & 2.34 & 2.34 & 4.69 & 1.87 & 2.34 & 4.69 \\ 
\end{tabular}
\end{table} 

\section{Results}
\label{sec:results}

In the experimental evaluation, we first compare three search strategies (a hill-climbing algorithm, co-evolution with a single LLM call per template, and co-evolution with multiple LLM calls per template) utilizing the same total number of LLM calls to demonstrate that the proposed co-evolutionary scheme yields the most promising results. Secondly, we perform circuit approximation for different error-area targets, using the maximum available LLM calls, to demonstrate that improved implementations of approximate 8-bit multipliers can be achieved for both WCE and MSE error metrics. 

\subsection{Search algorithms under a limited number of LLM calls}
\label{sec:res:comp}

\begin{figure}
    \centering
    \includegraphics[width=0.9\textwidth]{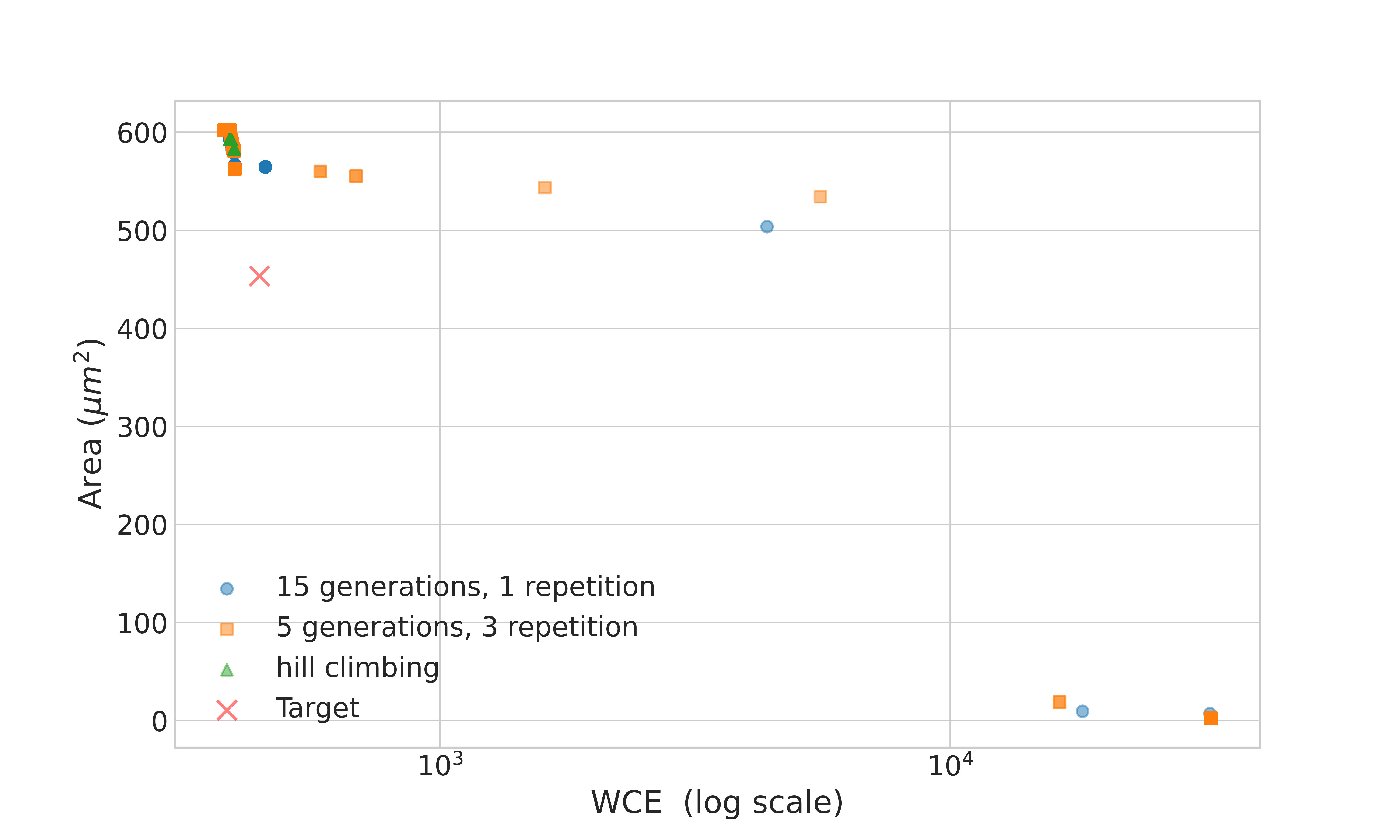}
    \caption{Final Pareto fronts obtained from 10 runs of HC, CoEA-3-5, and CoEA-1-15. The same initial population of circuits was used in the i-th run of CoEA algorithms. The target specification is depicted with a red cross.}
    \label{fig:pareto_run_comparison}
\end{figure}

We consider the number of calls for the bigger LLM (Qwen3-Coder-480B) per run (LLMpR) as a computational cost indicator. With LLMpR $= 450$, we compare the quality of the resulting Pareto fronts for 10 independent runs of three search algorithms targeting an 8-bit approximate multiplier with $T_E (WCE) =$ 21387 and $T_A =$ 453 $\mu m^2$. These target values are challenging, as seen in Fig.~\ref{fig:pareto_run_comparison}. They represent a hypothetical multiplier with roughly half the area compared to the exact multiplier. The algorithms considered in our study are:
\begin{enumerate}
\item \textbf{(HC)} A hill-climbing algorithm starts with a circuit selected from EvoApproxLib, which is 1.2$\times$--1.5$\times$ worse in both criteria than the target. It applies an LLM-driven mutation based on a single mutation template. The next parent is determined as the circuit with the smallest differences in both error and area from the target reached during the run. There is no co-evolution of templates. 
In each run, $g=$ LLMpR iterations are executed.

\item \textbf{(CoEA-1-15)} The proposed co-evolutionary algorithm is used; LLM is called once ($k=1$) per template, and $g = 15$ generations are produced.    

\item \textbf{(CoEA-3-5)} The proposed co-evolutionary algorithm is used; LLM is called 3 times ($k=3$) per template, and $g = 5$ generations are produced. As $k > 1$, $g$ has to be proportionally reduced to keep LLMpR constant.
\end{enumerate}

\begin{figure}
    \centering
    \includegraphics[width=0.7\textwidth]{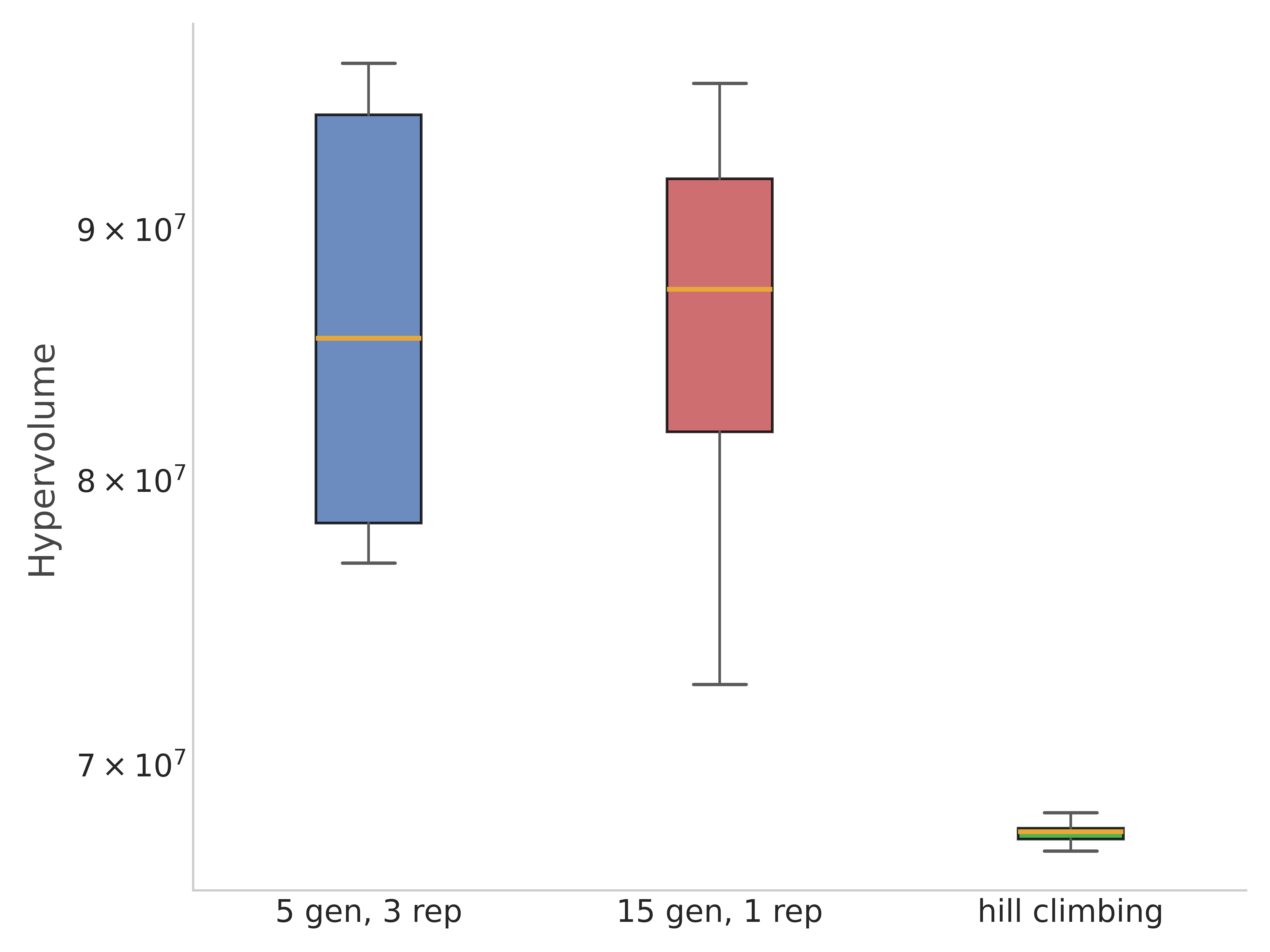}
    \caption{The hypervolume indicator computed from the final Pareto front constructed from 10 independent runs of HC, 6CoEA-3-5, and CoEA-1-15.}
    \label{fig:hv_boxplots}
\end{figure}

Fig.~\ref{fig:pareto_run_comparison} compares the final Pareto fronts from all runs of the algorithms considered. HC produces the least competitive results and tends to converge to a few almost identical solutions. The use of multiple LLM runs per template is a key feature of the proposed CoEA. With $k=3$, CoEA produced the highest number of unique solutions on the Pareto front; see the orange squares in Fig.~\ref{fig:pareto_run_comparison}. Under this setup, the search is rather exploratory as many solutions ``far'' from the specification are reached. Fig.~\ref{fig:hv_boxplots} analyzes the hypervolume indicators computed for each final Pareto front. A hypervolume is the dominated area in the objective space with respect to a reference point defined as the worst observed values increased by 1\%. The HC algorithm is clearly outperformed by CoEAs.

\begin{figure}
    \centering
    \includegraphics[width=0.9\textwidth]{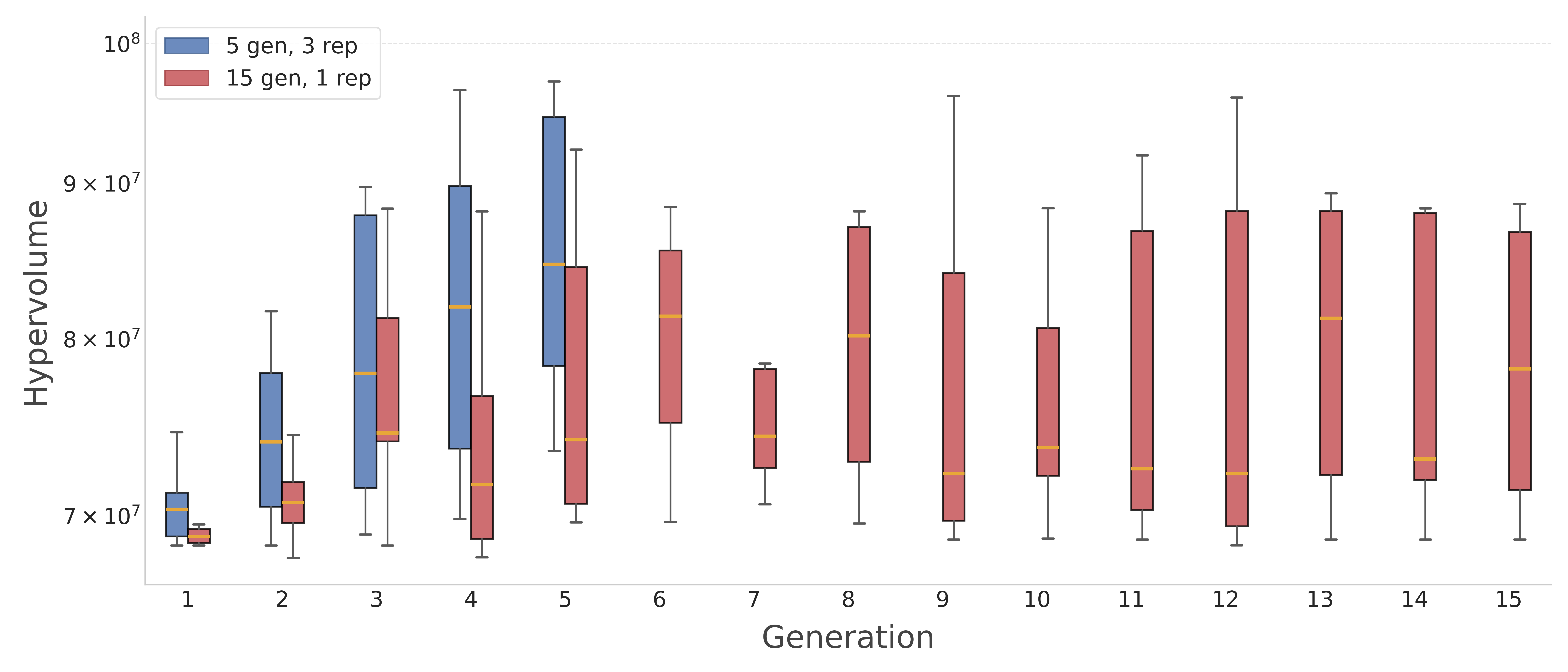}
    \caption{The progress of evolution of CoEA-3-5 and CoEA-1-15, where the hypervolume indicator is represented as a box plot of the best solutions (in 10 runs) reached in a given generation.}
    \label{fig:hv_boxplots:gen}
\end{figure}

The progress of evolution is visualized in Fig.~\ref{fig:hv_boxplots:gen}, where the hypervolume indicator is represented using a box plot of the best solutions (in 10 runs) reached in a given generation. It is important to note that CoEA-3-5 is executed for only 5 generations to ensure that both CoEA-3-5 and CoEA-1-15 use the same number of LLM calls.
Based on visual comparison of Fig.~\ref{fig:pareto_run_comparison}, ~\ref{fig:hv_boxplots}, and ~\ref{fig:hv_boxplots:gen}, CoEA-3-5 seems to perform better than  CoEA-1-15. However, a detailed statistical evaluation reported in Table~\ref{tab:mann_whitney} reveals that under the Mann–Whitney U test ($\alpha = 0.05$), no statistically significant differences are observed between CoEA-3-5 and CoEA-1-15, indicating comparable performance. Interestingly, CoEA-3-5 outperforms CoEA-1-15 if CoEA-1-15 uses slightly fewer resources (14 generations only). The average execution time of CoEA-3-5 is slightly shorter than that of CoEA-1-15 (207 min. vs 223 min).

\begin{table}[h]
\centering
\caption{Results of the Mann-Whitney U test for hypervolume comparison across experimental groups.}
\label{tab:mann_whitney}
\begin{tabular}{l|l|c|c|c}
\textbf{Comparison} & \textbf{Groups} & \textbf{U} & \textbf{p-value} & \textbf{Significant} \\
\hline\hline
\multicolumn{5}{l}{\textit{Overall hypervolume (Fig. 3)}} \\
\hline
CoEA-3-5 vs CoEA-1-15  & overall & 52.00 & 0.9097 & No \\
CoEA-3-5 vs hill climbing   & overall & 100.00 & 0.0002 & Yes*** \\
CoEA-1-15 vs hill climbing  & overall & 100.00 & 0.0002 & Yes*** \\
\hline\hline
\multicolumn{5}{l}{\textit{Final generations comparison (Fig. 4)}} \\
\hline
CoEA-3-5 (gen 5) vs CoEA-1-15 (gen 15) & per-gen & 75.00 & 0.0638 & No \\
CoEA-3-5 (gen 5) vs CoEA-1-15 rep (gen 14) & per-gen & 80.00 & 0.0256 & Yes* \\

\end{tabular}
\end{table}

\subsection{Long runs}
\label{sec:res:long}

\begin{figure}
    \centering
    \includegraphics[width=0.99\textwidth]{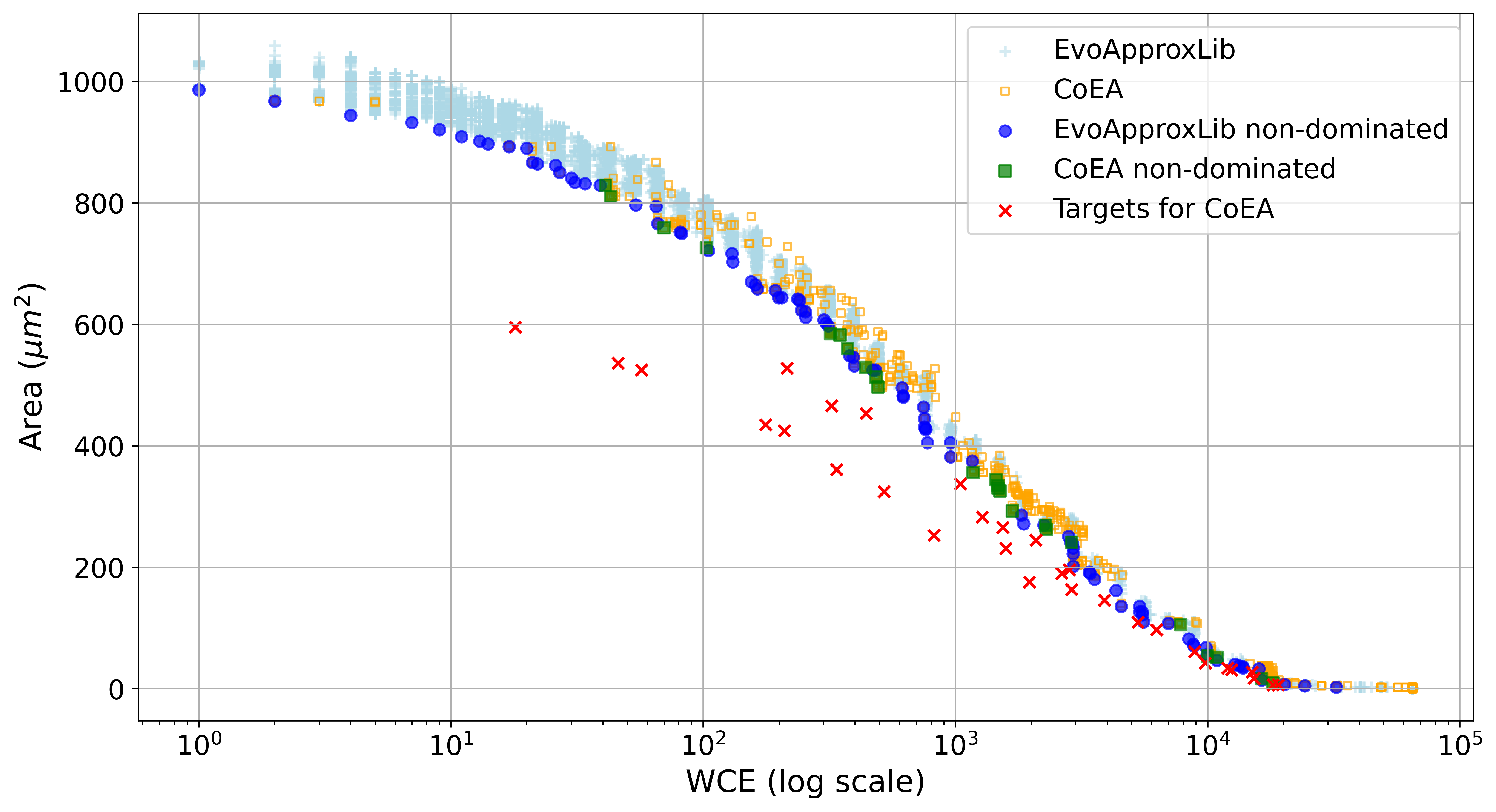}
    \caption{The \emph{WCE-Area} trade-offs for 8-bit approximate multipliers generated by the proposed CoEA or taken from EvoApproxLib. The united global Pareto front contains non-dominated solutions generated by CoEA (green square) and taken from EvoApproxLib (blue dots). The remaining points are dominated. The target specifications provided to CoEA are depicted with red crosses.}
    \label{fig:long_runs:wce}
\end{figure}

Based on the results reported in the previous section, we used CoEA with parameters specified in Section~\ref{sec:setup}, $k=3$ and $g=27$ to design approximate 8-bit multipliers for 32 target error-area pairs (see red crosses in Fig.~\ref{fig:long_runs:wce}). A single run, executed for each target pair, produced approximate multipliers that occupy a small local Pareto front. All global Pareto optimal designs produced by CoEA are depicted with green squares, and the Pareto optimal designs filtered from EvoApproxLib according to the WCE and area are shown as blue dots in Fig.~ \ref{fig:long_runs:wce}.

In 70 cases, approximate multipliers produced by CoEA show new trade-offs between WCE and area that are not present in EvoApproxLib; 24 of them are on the global Pareto front of all 8-bit multipliers. Note that the version of EvoApproxLib we employed for this comparison contains 24\,912 unsigned 8-bit approximate multipliers. The light blue plus symbols and the open orange squares in Fig.~\ref{fig:long_runs:wce} denote multipliers that are dominated by other designs.

\begin{figure}
    \centering
    \includegraphics[width=0.99\textwidth]{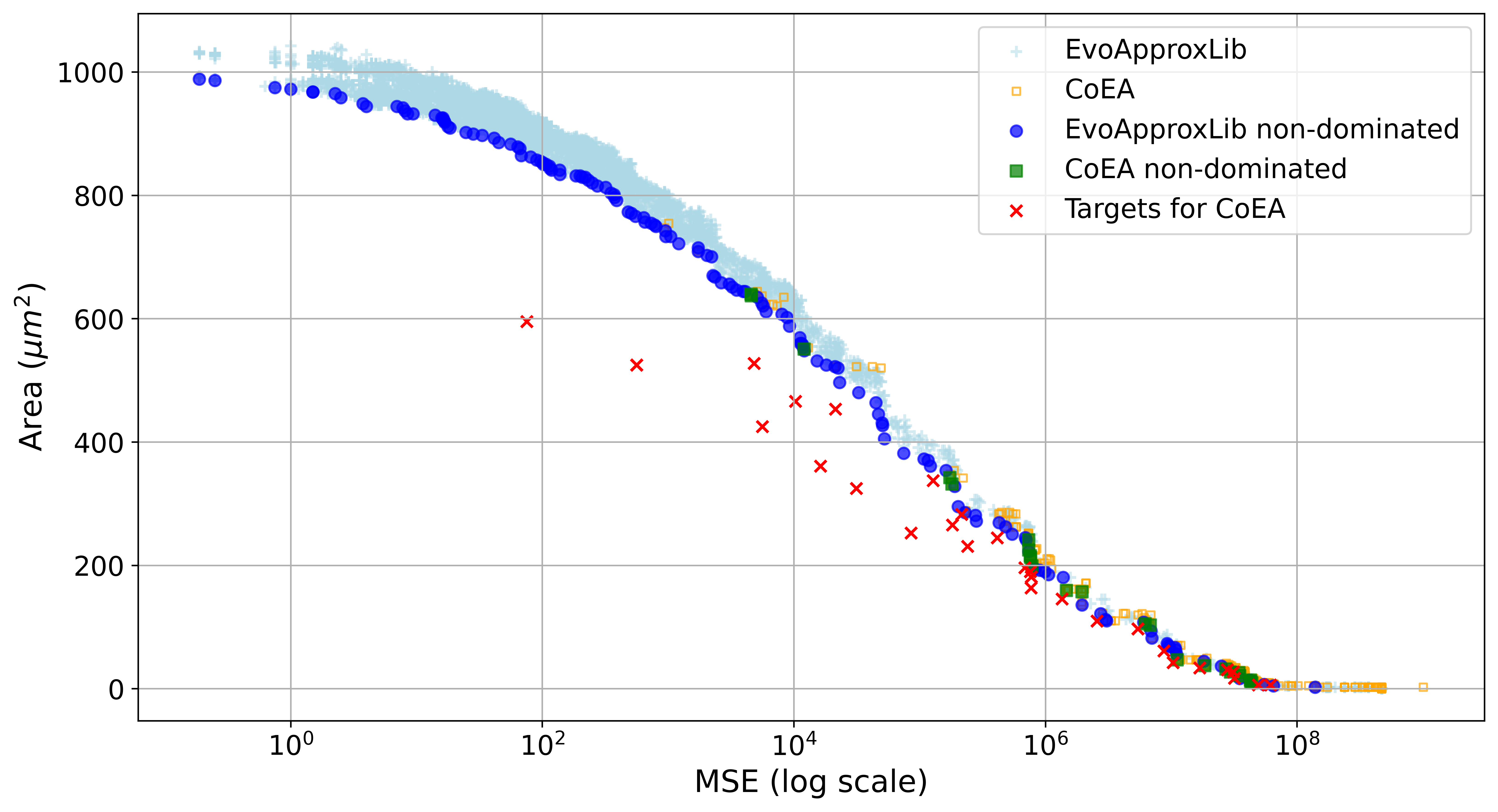}
    \caption{The \emph{MSE-Area} trade-offs for 8-bit approximate multipliers generated by the proposed CoEA or taken from EvoApproxLib. The united global Pareto front contains non-dominated solutions generated by CoEA (green square) and taken from EvoApproxLib (blue dots). The remaining points are dominated. The target specifications provided to CoEA are depicted with red crosses.}
    \label{fig:long_runs:mse}
\end{figure}

We changed the target error metric to MSE and repeated the experiment. Fig.~\ref{fig:long_runs:mse} shows that CoEA can deliver competitive approximate multipliers; in this case, 63 implementations show trade-offs unseen in EvoApproxLib, and 23 out of them are on the global Pareto front.

\subsection{Time requirements}

A single call to the LLM (Qwen3-Coder-480B) to generate a new circuit from existing ones takes about 10.0 seconds on average. The smaller model (GPT-OSS-120B) responds to a template-modification prompt in approximately 5.3 seconds.
By contrast, evaluating the error of a single candidate circuit takes only about 0.5 seconds on average, indicating that LLM inference dominates the overall runtime. Since the models were accessed through the shared computational infrastructure e-infra.cz, their inference times were outside our control.

Mrazek et al.~\cite{Mrazek:2020:approxMultipliersForCNN} report that a single CGP-based circuit approximation for one target error consumes up to $10^6$ generations (for a population size of 4), which is from 7.8 to 147.7 min on a 2.4 GHz Intel Xeon CPU, depending on the target error. The LLM-based approach requires one to two orders of magnitude fewer circuit evaluations.

\section{Conclusions}
\label{sec:concl}

We introduced a co-evolutionary design method that generates sequences of prompts, allowing a standard large language model to produce competitive approximate 8-bit multipliers without requiring prior model training. The method yielded designs with improved trade-offs compared to EvoApproxLib, achieving approximate multipliers with new WCE–area trade-offs (24 instances) and MSE–area trade-offs (23 instances). Notably, within hundreds to thousands of LLM calls, the approach produced highly optimized circuits that are complex (hundreds of gates) and whose optimization is considered a difficult combinatorial optimization problem.

Our experiments demonstrate that the co-evolution of circuits and prompt templates plays an important role in achieving high-quality error–area trade-offs. Alternative strategies such as extended hill-climbing or co-evolution with a restricted number of LLM calls per template did not achieve comparable performance.

Future work will include a detailed statistical analysis of long-run behavior, a thorough comparison of alternative CoEA and LLM configurations, further optimization of the core algorithm to reduce execution time, extension of the method to additional types of approximate circuits, and exploration of further design criteria.

\section*{Acknowledgements}
This work was supported by the Czech Science Foundation project 24-10990S and the Ministry of Education, Youth and Sports of the Czech Republic through the e-INFRA CZ (ID:90254) project.

\bibliographystyle{splncs04}
\bibliography{coevollm}

\end{document}